# Predicting Oral Disintegrating Tablet Formulations by Neural Network Techniques

Run Han[1], Yilong Yang[1,2], Xiaoshan Li[2], Defang Ouyang[1*]

[1]State Key Laboratory of Quality Research in Chinese Medicine, Institute of Chinese Medical Sciences (ICMS), University of Macau, Macau, China

[2]Department of Computer and Information Science, Faculty of Science and Technology, University of Macau, Macau, China

[*]Corresponding author: Defang Ouyang; Email: defangouyang@umac.mo;

**Abstract**

Oral Disintegrating Tablets (ODTs) is a novel dosage form that can be dissolved on the tongue within 3min or less especially for geriatric and pediatric patients. Current ODT formulation studies usually rely on the personal experience of pharmaceutical experts and trial-and-error in the laboratory, which is inefficient and time-consuming. The aim of current research was to establish the prediction model of ODT formulations with direct compression process by Artificial Neural Network (ANN) and Deep Neural Network (DNN) techniques. 145 formulation data were extracted from Web of Science. All data sets were divided into three parts: training set (105 data), validation set (20) and testing set (20). ANN and DNN were compared for the prediction of the disintegrating time. The accuracy of the ANN model has reached 85.60%, 80.00% and 75.00% on the training set, validation set and testing set respectively, whereas that of the DNN model was 85.60%, 85.00% and 80.00%, respectively. Compared with the ANN, DNN showed the better prediction for ODT formulations. It is the first time that deep neural network with the improved dataset selection algorithm is applied to formulation prediction on small data. The proposed predictive approach could evaluate the critical parameters about quality control of formulation, and guide research and process development. The implementation of this prediction model could effectively reduce drug product development timeline and material usage, and proactively facilitate the development of a robust drug product.

# 1. Introduction

Oral dosage forms are always the most widely used dosage form because of their convenience of self-administration, good stability, accurate dosing and easy manufacturing[1]. However, swallowing difficulty of the pediatric or geriatric patient is a big concern for conventional tablets. Dysphagia is observed in about 35% of the general population among all age groups, as well as in up to 40% of the elder population and 18-22% of all patients in long-term care facilities[2]. To overcome the difficulty in swallowing, oral disintegrating tablets (ODTs) have been developed since the 1990s[3, 4]. ODTs are designed to be dissolved on the tongue rather than swallowed whole as conventional tablets [5, 6]. The disintegrating time of ODTs is within 3 min or less in the saliva without the intake of water [7, 8]. In recent years, there is the growing demand about good ODT formulations with new disintegrants and convenient preparation methods. There are three major techniques which are widely used for ODT manufacture: freeze drying, tablet molding, tablet compression [9, 10]. Comparing with many other preparation methods, direct compression is most widely used because of its most effective and simplest process[11]. The formulations of ODTs with direct compression method usually contain the filler, binder, disintegrant, lubricant and solubilizer[12]. Therefore, formulation design of ODTs is critical to minimize the disintegrating time with good tablet quality.

Current pharmaceutical formulation development usually depends on experimental trial-and-error by personal experiences of formulation scientists, which is inefficient and time-consuming. To improve the efficiency of formulation screening, the SeDeM diagram expert system was developed to optimize formulations[13]. SeDeM diagram expert system was able to evaluate the influence of every excipient on the final formulation for direct compression based on the experimental study and quantitative characterization parameters[14]. Then this expert system considered the type of excipients and physicochemical properties to output a recommended formulation. Moreover, the mathematical analysis of SeDeM was able to recommend not only formulation components but also the optimal ratios of excipients [14, 15]. Firstly, 43 excipients were investigated the suitability for direct compression, especially the compressibility of disintegrants. According to the ICHQ8, the suitability was described as these parameters: bulk density, tapped density, inter-particle porosity, Carr index, cohesion index, Hausner ratio, angle of repose, powder flow, loss on drying, hygroscopicity, particle size and homogeneity index. The

SeDeM system could show the profile of every excipient and evaluate how suitable it can be used for direction compression[12]. According to the predicted result and combining with the experimental study, 8 excipients with the better properties were chosen to make a comparison using the new expert system. Compared with the old system, the new system could quantify the compressibility index of every excipient with the higher precision[16]. For example, ibuprofen ODT formulations were investigated with the suitability of 21 excipients and obtained the final SeDeM diagram with 12 parameters[17]. Current SeDeM method just focused on the recommended formulation, but it cannot quantitatively predict the disintegrating time of ODT formulations. With the challenge of pharmaceutical research, we need to establish a prediction method to assist experts evaluate the performance of ODT formulations.

The neural network is a wonderful biologically-inspired model that learn from observational data. That is an artificial network with seriously connected units by simulating the neural structure of the brain[18]. Neural network has been applied to solve problems in many fields, such as voice recognition and computer vision. Artificial neural network and deep neural network are two widely used neural networks, as shown in Fig. 1&2 [19]. ANN is a simple neuron network with only one hidden layer, while DNN is a more powerful technique with many complex layers to reach the high-level data representation. In pharmacology and bioinformatics research, ANN also has been used over two decades, included prediction of protein secondary structure and quantitative structure-activity relationship[20]. As the pharmaceutical research, the prediction models were developed for break force and disintegration of tablet formulation by ANN, genetic algorithm, support vector machine and random forest approaches [21]. Another ANN example was quantitative structure activity relationships (QSAR) of antibacterial activity study[22, 23]. DNN is a type of representation learning with multiple levels of neural networks. Unlike the traditional ANN with manual feature extraction, deep-learning can automatically extract feature even transform low-level representation to more abstract level without any feature extractor [24]. Moreover, deep-learning is more sensitive to irrelevant and particular minute variations with complicated parameters of the network, which could reach higher accuracy rather than the conventional machine learning algorithms [19]. In recent years, DNN has been applied in pharmacy research, such as drug design, drug-induced liver injury and virtual screening[25]. In most cases, deep-learning could generate a novel and complex system to represent various objects through molecular descriptor so that it would be very helpful for drug discovery and prediction[26].

Junshui Ma et al. extracted data from internal Merck data and included on-target and absorption, distribution, metabolism, excretion (ADME), each molecular was described as serious features. Finally, they use deep neural nets to evaluate QSAR and the result was better than random forest commonly used[27].

The aim of current research was to establish the quantitative prediction model of the disintegrating time of ODT formulations with direct compression process by ANN or DNN.

## 2. Methodology

*2.1. Data Extraction*

Formulation data collection was the foundation of building the prediction model. To ensure the data reliability, the keyword search strategy was used in Web of Science database. The synonym strings of keywords were used, such as "oral" + "disintegrating" + "tablets" with 461 results, "fast" + "disintegrating" + "tablets" with 407 results, "rapidly" + "disintegrating" + "tablets" with 266 results, and "oral" + "dispersible" + "tablets" with 84, respectively. Among these results, only research articles were selected for further data extraction. After the manual screening, 145 direct compressed ODT formulations with the disintegrating time were extracted including 23 active pharmacological ingredients (API) groups for our model, as shown in Table 1. All APIs were described as ten molecular parameters, including molecular weight, XLogP3, hydrogen bond donor count, hydrogen bond acceptor count, rotatable bond count, topological polar surface area, heavy atom count, complexity and logS. According to the function of excipients, all excipients were divided into five categories: filler, binder, disintegrant, lubricant, and solubilizer. Each type of excipients was individually coded for further training. The formulation data included API molecular descriptors and its amount, the type of encoded excipients and its amount, manufacture parameters (e.g. the hardness, friability, thickness and tablet diameter) and the disintegrating time of each formulation.

*2.2.Dataset Classification: Training set, Validation set and Testing set*

To ensure good prediction ability of computational model, especially in the small amount of pharmaceutical data, the dataset should be carefully divided into three parts, including training set, validation set and testing set. The three datasets strategy is an effective way to test the accuracy on new data out of our datasets. In details, the training set is for training model and the validation set is used for adjusting the parameters and finding the best model, while testing set shows the prediction accuracy on real unknown data from the datasets, as shown in Fig. 3. Therefore, how to select data for three datasets appropriately is the key step. Compared with random selection, manual selection and maximum dissimilarity algorithm selection, the improved maximum dissimilarity algorithm (MD-FIS) is the best choice. MD-FIS is based on the maximum dissimilarity algorithm considering with small group data in the whole dataset, it will avoid selecting data mostly from small group and ensure the representation of validation and test set.

*2.3. Hyperparameters of Artificial Neutral Network and Deep Neural Network*

The prediction model for ODTs was trained by ANN and DNN, respectively. In the training process, all data are normalized and then divided into three sets with our previous proposed *MD-FIS* selection algorithm in R language. For ANN and DNN network, Deeplearning4j machine learning framework (*https://deeplearning4j.org/)* was used to train prediction models. All the source code can be found on the website (*http://ml.mydreamy.net/pharmaceutics/ODT.html*). The ANN model in Figure. 1. with termination condition at 15000 epochs and hidden nodes is 200. The deep-learning process in Figure. 2. use full-connected deep feedforward networks including ten layers with 2000 epochs. This neural network contains 50 hidden nodes on each layer. All networks choose *tanh* as the activation function except the last layer with *sigmoid* activation function. Learning rate is set to 0.01. Batch gradient descent with the 0.8 *momentum* is used for training the networks.

Note that epoch indicates how many times the dataset is used for training. Feed-forward network means that the output of the network is computed layer-by-layer from one-direction without any inside loop. Learning rate impacts how fast the network will be convergent. Batch gradient descent is a training strategy to use all dataset to train the model at each time. Momentum indicates how much the speed will be kept in each training step.

*2.4. Pharmaceutical Evaluation Criterion*

European Pharmacopeia defined that ODT could disintegrate within 3 min in the mouth before being swallowed. In all our formulation data, the disintegrating time ranges from 0 sec to 100 sec. Usually, the successful prediction in pharmaceutics is that absolute error is less than 10%. Thus, a good model is that the prediction deviation of the disintegrating time is not more than 10sec. The accuracy of prediction disintegrating time is the percentage of successful prediction to total predictions:

$$accuracy_{PDT} = \frac{Number(|f' - f| \leq 10)}{AllPredictions}$$

Where, $f'$ is the prediction value, $f$ is the label (real) value. All predictions are the number of predicted data.

## 3. Results and discussion

Fig. 4 showed the label (true) value and predictive value of disintegrating time on ANN model (A. training set; B. validation set; C. testing set), while indicated the true value and predictive value of disintegrating time on DNN model (D. training set; E. validation set; F. testing set). As shown in Figure. 4, the training set and validation set of both ANN and DNN showed good results. As Table 2 shows, the predictive accuracy of ANN model is 85.60% on training set and 80.00% on validation set, while the DNN model is 85.60% and 85.00%, respectively. However, the testing set of ANN with only 75.00% accuracy is lower than that of DDN (80.00%), which indicated that DNN is able to significantly better predict real unknown data than ANN.

As the result shows, ANN is an efficient network for training prediction model within the adjustment of validation set, reaching a high accuracy on training set and validation set. However, when predicting real unknown data, the accuracy of testing set dropped significantly, which is called overfitting in machine learning. DNN performs well in all three data sets with over 80% accuracy and predicted stably with average value, which is more capable of establishing a better prediction model for ODT than ANN.

When analyzing the different network structure between ANN and DNN, ANN just includes one hidden layer, while DNN includes ten layers with 2000 epochs and each layer contains 50 hidden nodes. Thus, DNN could extract the feature of data with higher level and give a more accurate predictive result. It is unsurprised that DNN, as an innovative and effective technique for pharmaceutical research, can provide a higher accuracy prediction about disintegrating time than ANN. Thus, the desired DNN with the proposed MD-FIS selection algorithm can be used to achieve good predictive results on pharmaceutical formulations with small data.

In order to ensure a satisfied prediction accuracy, two key factors are to be considered: data and algorithm. The first issue is the reliable data in pharmaceutical research. Deep-learning attempts to learn these characteristics to make better representations and create models from reliable data. Thus, data extraction is a critical step. In current research, reliable formulation data set were manually extracted and labeled from the research articles of Web of Science by experienced pharmaceutical scientists.

On the other hand, small data in pharmaceutical research is the key issue to be solved. Although there are many DNN examples about imaging recognition, natural language processing and auto-mobile car, it is still very few pharmaceutical researches about deep-learning. Usually speaking, deep learning methods require a large amount of data for training. This is not a problem in other fields which have the big data source. However, this is a big challenge for the pharmaceutical researches due to the experimental limitation. Thus, the most important problem is: how to train a good prediction model on small data with high-dimensions input space? For example, the formulation data of ODTs includes the chemical and physical properties of APIs, multiple excipients with various ratios and four tablets characteristic parameters. In our 145-formulation data, it was found that near half of APIs groups' size is less than 3 (small API group). Therefore, the splitting strategy of data set is critical for model establish.  Firstly, 20 representative testing set were picked up from the whole dataset by pharmaceutical scientists. As for training set and validation set selection, before using automatic selection algorithm, manual selection approach was adopted to ensure the appropriate selection of these two data sets. However, the manual selection needs experts with strong background knowledge, which is time-consuming and non-standardized. When trying the random selection method, the data from small API groups with no representation was easily selected.  Thus, the improved maximum dissimilarity algorithm (MD-FIS) is developed to select training set and validation set. MD-FIS is based on the maximum dissimilarity algorithm with the small group filter, representative initial set selection algorithm and new selection cost function. In the MD-FIS process, the data go through a filter to get rid of the data from the small API groups, then the MD-FIS randomly get the initial data sets, compute each distance from the initial data set to the corresponding remaining data, the minimum distance data are chosen as the final initial set. The final initial set and remaining data are the input to the dissimilarity algorithm with new selection cost function. The selected data is the validation set, while the remaining data is used as the training set. Because of the small group filter, the validation set from the general groups could represent the feature of whole data set.

The second important issue is the selection of network algorithm. As deep convolutional networks inspired from visual neuroscience usually achieve a good result for processing images, video, speech and audio[28]. Recurrent neural networks contained history information of the sequence have brought the breakthrough in sequential data such as text and speech[29]. Our pharmaceutics data only includes properties of API, excipients with its amount and tablet parameters. There is no chronic relationship between each data. Our target is to predict the disintegrating time. Hence, compared with the deep convolutional networks and recurrent neural networks, the full-connected deep feedforward networks should be the best choice for the proposed problem. The challenge about deep feedforward network is computing too many parameters and vanishing the gradient. The results show that the satisfied accuracy could be reached by DNN. The deep learning method with the proposed data selection algorithms and pharmaceutics evaluation criterion can reach the desired models, which satisfy the accuracy requirements in the pharmaceutics. This deep-learning approach could save a lot of time, manpower and material resource for formulation development of ODTs. This will greatly benefit the formulation design in pharmaceutical research.

Although DNN has reached the expected prediction accuracy on small pharmaceutical data sets, the mechanism of DNN is still a black box, and it is difficult to explain the mapping procedure from the input layer to the output layer. For example, it is unclear how each formulation component contributes to the disintegrating time. Moreover, current model cannot be directly applied to another evaluation parameters of formulations. Current prediction model for ODTs is just the first step in intelligent research for formulation development. Further research in intelligent formulation systems is underway in our laboratory.

## 4. Conclusions

The traditional "trial-and-error" method for formulation development has existed hundreds of years, which always cost a large amount of time, financial and human resources. Oral disintegrating tablets is a novel and important formulation form in recent years because of its convenience and good disintegration ability. Current research developed the DNN with MD-FIS select algorithm to establish a good prediction model for the disintegrating time of ODT formulations. On the other hand, this research is also a good example for deep-learning on small data. The proposed predictive approach not only contains formulation information of ODTs, but consider with the influence of tablet characteristic parameters, it could evaluate the critical parameters about quality control of formulation, and guide formulation research and process development. This deep-learning model could also be applied to other dosage forms and more fields in pharmaceutical research. The implementation of this prediction model could effectively reduce drug product development timeline and material usage, and proactively facilitate the development of a robust drug product.

## Acknowledgments

Current research is financially supported by the University of Macau Research Grant (MYRG2016-00038-ICMS-QRCM & MYRG2016-00040-ICMS-QRCM), Macau Science and Technology Development Fund (FDCT) (Grant No. 103/2015/A3) and the National Natural Science Foundation of China (Grant No. 61562011).

# Figure and Table legends

Fig. 1. The network structure of ANN.

Fig. 2. The network structure of DNN.

Fig. 3. The flowchart of training model.

Fig. 4. The true value and predictive value on dataset: (A) the true value and predictive value of training set, (B) validation set and (C) testing set on ANN model; (D) the true value and predictive value of training set, (E) validation set and (F) testing set on DNN model.

Table 1 The formulation data of ODTs

Table 2 The accuracies of OFDT on training, testing, and final testing sets

# Figures:

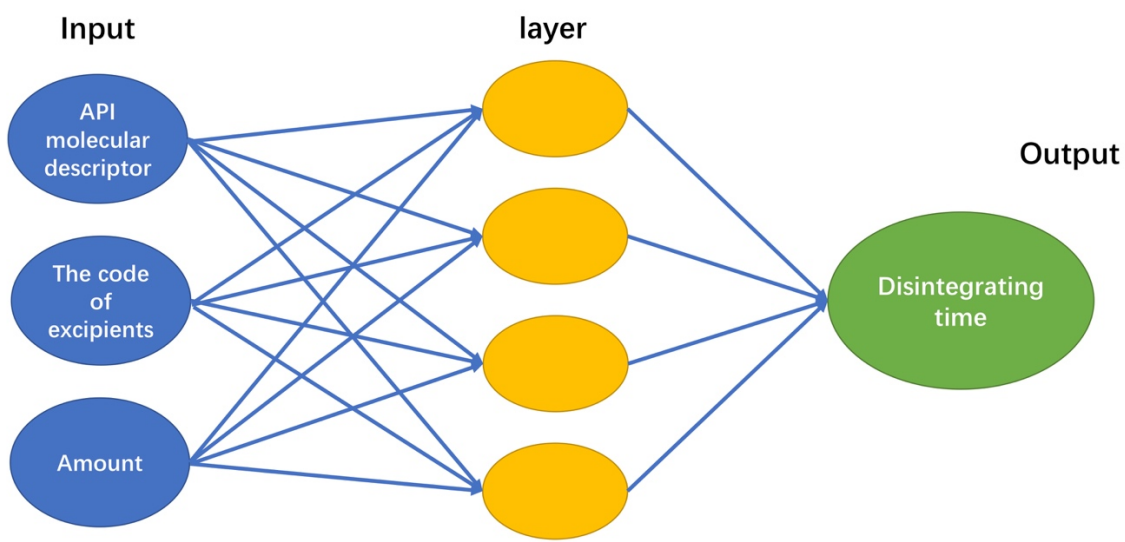

Fig. 1. The network structure of ANN

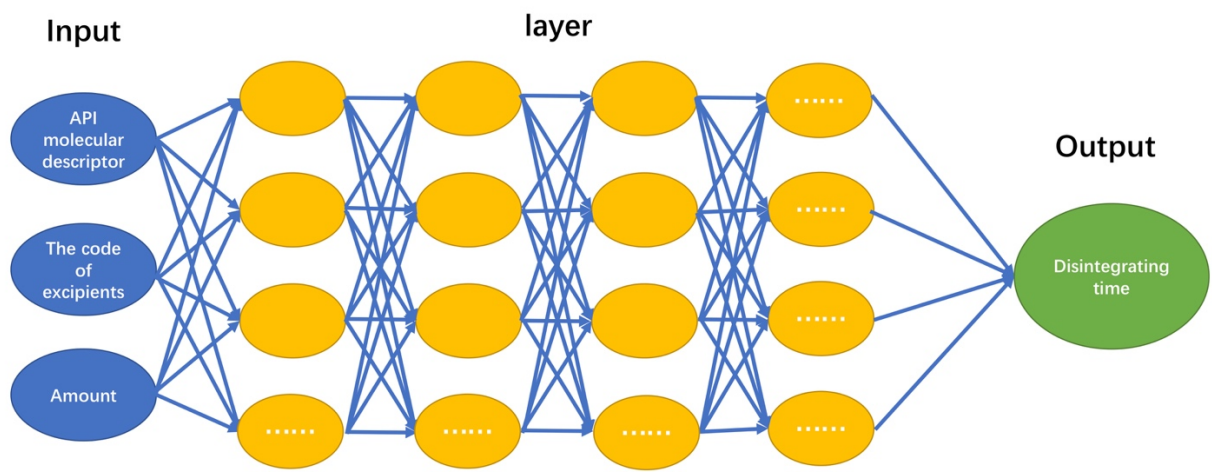

Fig. 2. The network structure of DNN

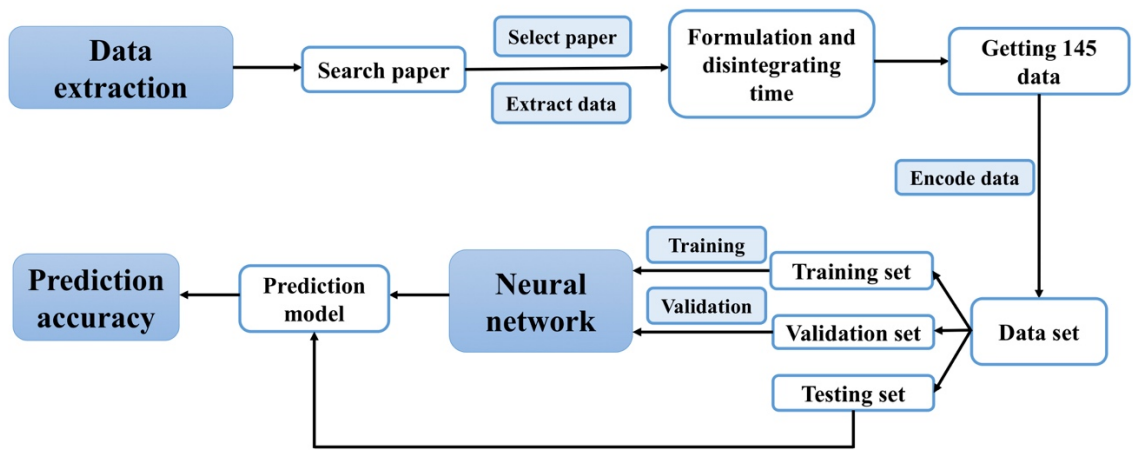

Fig. 3. The flowchart of establishing model

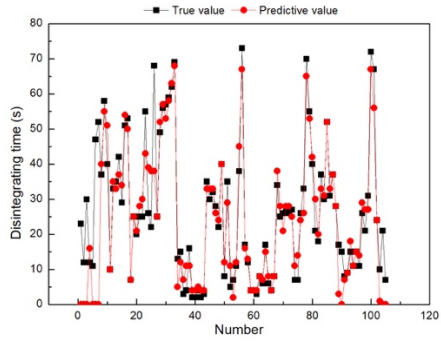
A

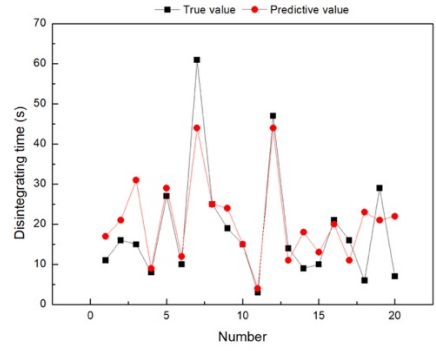
B

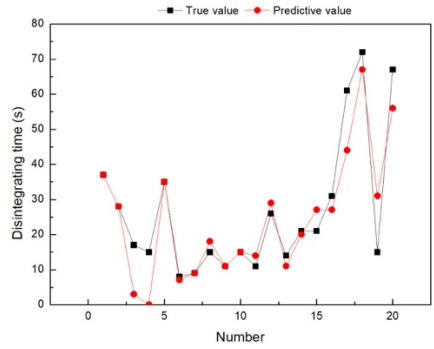
C

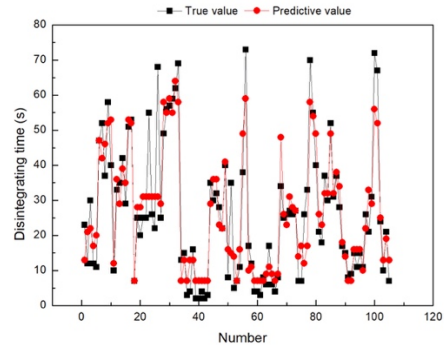
D

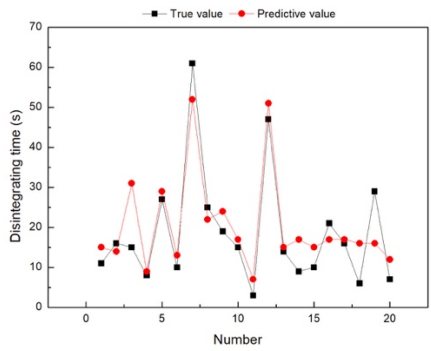
E

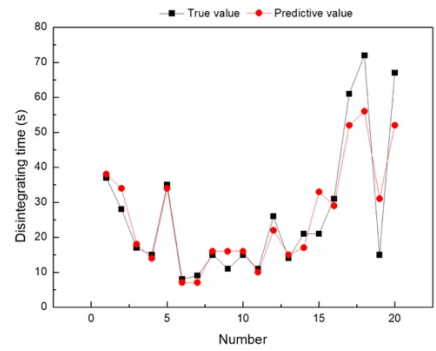
F

Fig. 4. The true value and predictive value on dataset: (A) the true value and predictive value of training set and (B) validation set and (C) testing set on ANN model. (D) the true value and predictive value of training set and (E) validation set and (F) testing set on DNN model.

**Tables:**

Table 1 The formulation data of ODTs

| API | Dose (mg) | Filler | Dose (mg) | Filler | Dose (mg) | Binder | Dose (mg) | Disintegrant | Dose (mg) | Disintegrant | Dose (mg) | Lubricant | Dose (mg) | Lubricant | Dose (mg) | Solubilzer | Dose (mg) | Hardness (N) | Friability (%) | Thickness (mm) | Punch (mm) | Disintegration time (sec) |
|---|---|---|---|---|---|---|---|---|---|---|---|---|---|---|---|---|---|---|---|---|---|---|
| Mirtazapine | 45 | Mannitol | 285 | MCC | 0 | PVP | 195 | CC-Na | 25 | | | Aerosil | 0 | Mg stearate | 10 | | | 53 | 0.56 | 4.76 | | 30 |
| Mirtazapine | 45 | Mannitol | 264 | MCC | 0 | PVP | 195 | CC-Na | 25 | | | Aerosil | 0 | Mg stearate | 10 | | | 50 | 0.52 | 4.75 | | 24 |
| Hydrochlorothiazide | 50 | Sucralose | 133.6 | | | | | CC-Na | 8 | PVPP | 8 | Aerosil | 15 | Mg stearate | 4 | | | 45 | | | 8 | 10 |
| Hydrochlorothiazide | 50 | Sucralose | 0 | | | | | CC-Na | 8 | PVPP | 8 | Aerosil | 15 | Mg stearate | 4 | | | 45 | | | 8 | 21 |
| Paracetamol | 224.4 | Mannitol | 303.6 | | | | | CC-Na | 44.4 | | | Mg stearate | 3 | | | | | 28 | 2.06 | | 11 | 37 |
| Paracetamol | 224.4 | Mannitol | 303.6 | | | | | CC-Na | 36.6 | | | Mg stearate | 3 | | | | | 41 | 0.88 | | 11 | 58 |
| Paracetamol | 224.4 | Mannitol | 291.6 | | | | | CC-Na | 32.4 | | | Mg stearate | 3 | | | | | 48 | 0.56 | | 11 | 40 |
| Paracetamol | 224.4 | Mannitol | 291.6 | | | | | CC-Na | 28.6 | | | Mg stearate | 3 | | | | | 50 | 0.65 | | 11 | 67 |
| Paracetamol | 325 | MCC | 113 | | | | | CC-Na | 0 | CMS-Na | 40 | Mg stearate | 2 | | | | | 45 | 0.86 | | 11 | 37 |
| Paracetamol | 325 | MCC | 113 | | | | | CC-Na | 40 | CMS-Na | 20 | Mg stearate | 2 | | | | | 45 | 0.69 | | 11 | 52.33 |
| Famotidine | 20 | Mannitol | 71.76 | Lactose | 0 | L-HPC | 0 | CC-Na | 2.34 | CMS-Na | 0 | Mg stearate | 0.5 | | | | | 46 | 0.95 | | 7 | 22.91 |
| Famotidine | 20 | Mannitol | 0 | Lactose | 71.6 | L-HPC | 0 | CC-Na | 2.34 | CMS-Na | 0 | Mg stearate | 0.5 | | | | | 65 | 0.96 | | 7 | 11.69 |
| Famotidine | 20 | Mannitol | 0 | Lactose | 0 | L-HPC | 0 | CC-Na | 2.34 | CMS-Na | 0 | Mg stearate | 0.5 | | | | | 60 | 1.25 | | 7 | 14.63 |
| Famotidine | 20 | Mannitol | 75.66 | Lactose | 0 | L-HPC | 0 | CC-Na | 0 | CMS-Na | 6.24 | Mg stearate | 0.5 | | | | | 57 | 0.99 | | 7 | 17.19 |
| Famotidine | 20 | Mannitol | 0 | Lactose | 75.66 | L-HPC | 0 | CC-Na | 0 | CMS-Na | 6.24 | Mg stearate | 0.5 | | | | | 92 | 1.02 | | 7 | 30.27 |
| Famotidine | 20 | Mannitol | 0 | Lactose | 0 | L-HPC | 0 | CC-Na | 0 | CMS-Na | 6.24 | Mg stearate | 0.5 | | | | | 103 | 0.98 | | 7 | 12.48 |
| Famotidine | 20 | Mannitol | 66.3 | Lactose | 0 | L-HPC | 11.7 | CC-Na | 0 | CMS-Na | 0 | Mg stearate | 0.5 | | | | | 55 | 0.97 | | 7 | 11.42 |
| Famotidine | 20 | Mannitol | 0 | Lactose | 66.3 | L-HPC | 11.7 | CC-Na | 0 | CMS-Na | 0 | Mg stearate | 0.5 | | | | | 108 | 1.13 | | 7 | 47.25 |
| Famotidine | 20 | Mannitol | 0 | Lactose | 0 | L-HPC | 11.7 | CC-Na | 0 | CMS-Na | 0 | Mg stearate | 0.5 | | | | | 121 | 0.92 | | 7 | 52.21 |
| Acetaminophen | 325 | MCC | 133 | | | | | CC-Na | 20 | CMS-Na | 0 | Mg stearate | 2 | | | | | 45 | 0.86 | | 11.1 | 33 |
| Acetaminophen | 325 | MCC | 113 | | | | | CC-Na | 40 | CMS-Na | 0 | Mg stearate | 2 | | | | | 46 | 0.43 | | 11.1 | 35 |
| Acetaminophen | 325 | MCC | 113 | | | | | CC-Na | 20 | CMS-Na | 0 | Mg stearate | 2 | | | | | 45 | 0.76 | | 11.1 | 24 |

| API | Dose | Filler | | Filler2 | | Disint1 | | Disint2 | | Disint3 | | Lubricant | | Glidant | | Other | | R1 | R2 | R3 | R4 | R5 |
|---|---|---|---|---|---|---|---|---|---|---|---|---|---|---|---|---|---|---|---|---|---|---|
| Acetaminophen | 325 | MCC | 133 | | | | | CC-Na | 0 | CMS-Na | 20 | Mg stearate | 2 | | | | | 47 | 0.92 | | 11.1 | 42.33 |
| Acetaminophen | 325 | MCC | 113 | | | | | CC-Na | 0 | CMS-Na | 20 | Mg stearate | 2 | | | | | 50 | 0.95 | | 11.1 | 29 |
| Acetaminophen | 325 | MCC | 133 | | | | | CC-Na | 20 | CMS-Na | 20 | Mg stearate | 2 | | | | | 52 | 0.78 | | 11.1 | 51 |
| Acetaminophen | 325 | MCC | 113 | | | | | CC-Na | 20 | CMS-Na | 20 | Mg stearate | 2 | | | | | 46 | 0.67 | | 11.1 | 53.33 |
| Olanzapine | 11.8 | Mannitol | 41 | MCC | 61.45 | | | CC-Na | 14 | | | Mg stearate | 0.875 | Aerosil | 0.875 | 2-hydroxypropyl-β-cyclodextrin | 43 | 36 | 0.78 | 3.13 | 8 | 27 |
| Olanzapine | 11.8 | Mannitol | 41 | MCC | 59.7 | | | CC-Na | 15.75 | | | Mg stearate | 0.875 | Aerosil | 0.875 | 2-hydroxypropyl-β-cyclodextrin | 43 | 35 | 0.82 | 3.18 | 8 | 25 |
| Olanzapine | 11.8 | Mannitol | 41 | MCC | 57.95 | | | CC-Na | 17.5 | | | Mg stearate | 0.875 | Aerosil | 0.875 | 2-hydroxypropyl-β-cyclodextrin | 43 | 33 | 0.87 | 3.34 | 8 | 20 |
| Olanzapine | 11.8 | Mannitol | 41 | MCC | 70.2 | | | CC-Na | 5.25 | | | Mg stearate | 0.875 | Aerosil | 0.875 | 2-hydroxypropyl-β-cyclodextrin | 43 | 33 | 0.85 | 3.13 | 8 | 25 |
| Olanzapine | 11.8 | Mannitol | 41 | MCC | 68.45 | | | CC-Na | 7 | | | Mg stearate | 0.875 | Aerosil | 0.875 | 2-hydroxypropyl-β-cyclodextrin | 43 | 36 | 0.85 | 3.26 | 8 | 25 |
| Olanzapine | 11.8 | Mannitol | 41 | MCC | 70.2 | | | CC-Na | 0 | | | Mg stearate | 0.875 | Aerosil | 0.875 | 2-hydroxypropyl-β-cyclodextrin | 43 | 36 | 0.85 | 3.14 | 8 | 55 |
| Olanzapine | 11.8 | Mannitol | 41 | MCC | 68.45 | | | CC-Na | 0 | | | Mg stearate | 0.875 | Aerosil | 0.875 | 2-hydroxypropyl-β-cyclodextrin | 43 | 34 | 0.86 | 3.14 | 8 | 26 |
| Olanzapine | 11.8 | Mannitol | 41 | MCC | 70.2 | | | CC-Na | 5.25 | | | Mg stearate | 0.875 | Aerosil | 0.875 | 2-hydroxypropyl-β-cyclodextrin | 43 | 35 | 0.82 | 3.76 | 8 | 28 |
| Olanzapine | 11.8 | Mannitol | 41 | MCC | 68.45 | | | CC-Na | 7 | | | Mg stearate | 0.875 | Aerosil | 0.875 | 2-hydroxypropyl-β-cyclodextrin | 43 | 32 | 0.79 | 3.64 | 8 | 21 |
| Olanzapine | 11.8 | Mannitol | 41 | MCC | 66.7 | | | CC-Na | 8.75 | | | Mg stearate | 0.875 | Aerosil | 0.875 | 2-hydroxypropyl-β-cyclodextrin | 43 | 35 | 0.52 | 3.23 | 8 | 22 |
| Olanzapine | 11.8 | Mannitol | 41 | MCC | 70.2 | | | CC-Na | 5.25 | | | Mg stearate | 0.875 | Aerosil | 0.875 | 2-hydroxypropyl-β-cyclodextrin | 43 | 35 | 0.75 | 3.25 | 8 | 31 |
| Olanzapine | 11.8 | Mannitol | 41 | MCC | 68.45 | | | CC-Na | 7 | | | Mg stearate | 0.875 | Aerosil | 0.875 | 2-hydroxypropyl-β-cyclodextrin | 43 | 32 | 0.67 | 3.25 | 8 | 27 |
| Olanzapine | 11.8 | Mannitol | 41 | MCC | 66.7 | | | CC-Na | 8.75 | | | Mg stearate | 0.875 | Aerosil | 0.875 | 2-hydroxypropyl-β-cyclodextrin | 43 | 38 | 0.65 | 3.21 | 8 | 68 |
| Eslicarbazepine | 800 | Mannitol | 150 | MCC | 70.08 | | | CC-Na | 0 | PVPP | 40 | Mg stearate | 4 | | | β-cyclodextrin | 109.9 | 38 | 0.85 | 6.5 | 16 | 45.33 |
| Eslicarbazepine | 800 | Mannitol | 150 | MCC | 50.08 | | | CC-Na | 0 | PVPP | 60 | Mg stearate | 4 | | | β-cyclodextrin | 109.9 | 37 | 0.75 | 6.5 | 16 | 24.66 |
| Eslicarbazepine | 800 | Mannitol | 150 | MCC | 70.08 | | | CC-Na | 0 | PVPP | 0 | Mg stearate | 4 | | | β-cyclodextrin | 109.9 | 38 | 0.81 | 6.5 | 16 | 49.33 |
| Eslicarbazepine | 800 | Mannitol | 150 | MCC | 50.08 | | | CC-Na | 0 | PVPP | 0 | Mg stearate | 4 | | | β-cyclodextrin | 109.9 | 38 | 0.87 | 6.5 | 16 | 55.66 |
| Eslicarbazepine | 800 | Mannitol | 150 | MCC | 70.08 | | | CC-Na | 0 | PVPP | 0 | Mg stearate | 4 | | | β-cyclodextrin | 109.9 | 37 | 0.72 | 6.5 | 16 | 57.33 |
| Eslicarbazepine | 102 | Mannitol | 150 | MCC | 50.08 | | | CC-Na | 0 | PVPP | 60 | Mg stearate | 4 | | | β-cyclodextrin | 109.9 | 38 | 0.72 | 6.5 | 16 | 24.66 |
| Eslicarbazepine | 102 | Mannitol | 150 | MCC | 70.08 | | | CC-Na | 40 | PVPP | 0 | Mg stearate | 4 | | | β-cyclodextrin | 109.9 | 38 | 0.81 | 6.5 | 16 | 61.66 |
| Lornoxicam | 4 | Mannitol | 63.5 | MCC | 15 | L-HPC | 3 | CC-Na | 7.5 | | | Mg stearate | 1 | Aerosil | 1 | Cyclodextrin Methacrylate | 0 | 24 | 0.42 | 2.14 | 12 | 7.4 |
| Lornoxicam | 4 | Mannitol | 63.5 | MCC | 15 | L-HPC | 3 | CC-Na | 7.5 | | | Mg stearate | 1 | Aerosil | 1 | Cyclodextrin Methacrylate | 4 | 22 | 0.28 | 2.22 | 12 | 7.3 |

| API | Amount | Filler | Amount | Filler2 | Amount | Disint | Amount | Disint2 | Amount | Disint3 | Amount | Lubricant | Amount | Lubricant2 | Amount | Lubricant3 | Amount | Other | Amount | C1 | C2 | C3 | C4 | C5 |
|---|---|---|---|---|---|---|---|---|---|---|---|---|---|---|---|---|---|---|---|---|---|---|---|---|
| Lornoxicam | 4 | Mannitol | 63.5 | MCC | 15 | L-HPC | 3 | CC-Na | 7.5 | | | Mg stearate | 1 | Aerosil | 1 | | | Cyclodextrin Methacrylate | 12.21 | 23 | 0.36 | 2.21 | 12 | 7.4 |
| Meloxicam | 7.5 | Mannitol | 20 | MCC | 40 | | | PVPP | 10 | | | Mg stearate | 1 | | | | | | | 27 | 0.99 | 2.03 | 9.58 | 46.17 |
| Miconazole nitrate | 56.5 | Mannitol | 58 | MCC | 58 | HPMC | 4.7 | CC-Na | 4.8 | | | Mg stearate | 0 | SDS | 18 | | | | | 56 | 0.45 | 3.53 | 8 | 40 |
| Miconazole nitrate | 56.5 | Mannitol | 78 | MCC | 26 | HPMC | 4.7 | CC-Na | 0 | | | Mg stearate | 0 | SDS | 24 | | | | | 66 | 0.67 | 3.02 | 8 | 35 |
| Miconazole nitrate | 56.5 | Mannitol | 58 | MCC | 58 | HPMC | 4.7 | CC-Na | 14.4 | | | Mg stearate | 6 | SDS | 0 | | | | | 79 | 0.18 | 3.52 | 8 | 18 |
| Dextromethorphan | 15 | Mannitol | 10 | MCC | 25 | | | | | | | | | | | | | | | 37 | 0.76 | 3.64 | 9.53 | 21 |
| Dextromethorphan | 15 | Mannitol | 10 | MCC | 25 | | | | | | | | | | | | | | | 40 | 0.74 | 3.56 | 9.53 | 13.8 |
| Risperidone | 0.5 | Mannitol | 1.3 | MCC | 2.6 | PVP | 0 | CC-Na | 0 | CMS-Na | 0.5 | Aerosil | 50 | | | | | | | | | 1.58 | 3 | 14.83 |
| Risperidone | 0.5 | Mannitol | 1.3 | MCC | 2.6 | PVP | 0.5 | CC-Na | 0 | CMS-Na | 0 | Aerosil | 50 | | | | | | | | | 1.65 | 3 | 12.97 |
| Risperidone | 0.5 | Mannitol | 2.6 | MCC | 1.3 | PVP | 0 | CC-Na | 0.5 | CMS-Na | 0 | Aerosil | 50 | | | | | | | | | 1.65 | 3 | 2.99 |
| Risperidone | 0.5 | Mannitol | 2.6 | MCC | 1.3 | PVP | 0 | CC-Na | 0 | CMS-Na | 0.5 | Aerosil | 50 | | | | | | | | | 1.66 | 3 | 4.39 |
| Risperidone | 0.5 | Mannitol | 1.45 | MCC | 1.45 | PVP | 0 | CC-Na | 0 | CMS-Na | 0.5 | Aerosil | 50 | | | | | | | | | 1.64 | 3 | 15.91 |
| Risperidone | 0.5 | Mannitol | 2.6 | MCC | 1.3 | PVP | 0.5 | CC-Na | 0 | CMS-Na | 0 | Aerosil | 50 | | | | | | | | | 1.63 | 3 | 1.68 |
| Risperidone | 0.5 | Mannitol | 2.6 | MCC | 1.3 | PVP | 0.5 | CC-Na | 0 | CMS-Na | 0 | Aerosil | 50 | | | | | | | | | 1.77 | 3 | 2.19 |
| Risperidone | 0.5 | Mannitol | 1.3 | MCC | 2.6 | PVP | 0 | CC-Na | 0.5 | CMS-Na | 0 | Aerosil | 50 | | | | | | | | | 1.67 | 3 | 8.01 |
| Risperidone | 0.5 | Mannitol | 1.3 | MCC | 2.6 | PVP | 0.5 | CC-Na | 0 | CMS-Na | 0 | Aerosil | 50 | | | | | | | | | 1.61 | 3 | 3.93 |
| Risperidone | 0.5 | Mannitol | 1.45 | MCC | 1.45 | PVP | 0 | CC-Na | 0 | CMS-Na | 0.5 | Aerosil | 50 | | | | | | | | | 1.63 | 3 | 9.17 |
| Risperidone | 0.5 | Mannitol | 1.45 | MCC | 1.45 | PVP | 0.5 | CC-Na | 0 | CMS-Na | 0 | Aerosil | 50 | | | | | | | | | 1.65 | 3 | 2.41 |
| Risperidone | 0.5 | Mannitol | 1.45 | MCC | 1.45 | PVP | 0.5 | CC-Na | 0 | CMS-Na | 0 | Aerosil | 50 | | | | | | | | | 1.6 | 3 | 2.61 |
| Risperidone | 0.5 | Mannitol | 1.45 | MCC | 1.45 | PVP | 0.5 | CC-Na | 0 | CMS-Na | 0 | Aerosil | 50 | | | | | | | | | 1.63 | 3 | 2.81 |
| Granisetron | 50 | Mannitol | 20 | MCC | 55 | | | CC-Na | 0 | CMS-Na | 5 | Aerosil | 2 | Mg stearate | 1.5 | | | | | 35 | 0.2 | 4.38 | 6 | 35 |
| Granisetron | 50 | Mannitol | 20 | MCC | 52.5 | | | CC-Na | 0 | CMS-Na | 7.5 | Aerosil | 2 | Mg stearate | 1.5 | | | | | 40 | 0.13 | 4.31 | 6 | 30 |
| Granisetron | 50 | Mannitol | 20 | MCC | 55 | | | CC-Na | 5 | CMS-Na | 0 | Aerosil | 2 | Mg stearate | 1.5 | | | | | 45 | 0.14 | 4.39 | 6 | 32 |
| Granisetron | 50 | Mannitol | 20 | MCC | 52.5 | | | CC-Na | 7.5 | CMS-Na | 0 | Aerosil | 2 | Mg stearate | 1.5 | | | | | 35 | 0.13 | 4.37 | 6 | 28 |
| Granisetron | 50 | Mannitol | 20 | | 52.5 | | | CC-Na | 0 | CMS-Na | 0 | Aerosil | 2 | Mg stearate | 1.5 | | | | | 30 | 0.21 | 4.34 | 6 | 22 |
| Mefenamic | 100 | MCC | 81.75 | | | | | PVPP | 32.5 | | | Aerosil | 32.5 | Mg stearate | 3.25 | | | | | 18 | 0.92 | 4.1 | 12 | 25 |

| Drug | mg | Filler 1 | mg | Filler 2 | mg | Binder | mg | Disintegrant 1 | mg | Disintegrant 2 | mg | Glidant | mg | Lubricant | mg | Other | mg | Col1 | Col2 | Col3 | Col4 | Col5 |
|---|---|---|---|---|---|---|---|---|---|---|---|---|---|---|---|---|---|---|---|---|---|---|
| Mefenamic | 100 | MCC | 181.75 | | | | | PVPP | 32.5 | | | Aerosil | 32.5 | Mg stearate | 3.25 | | | 22 | 0.65 | 3.8 | 12 | 25 |
| Atorvastatin | 10 | Mannitol | 175 | | | L-HPC | 15 | CC-Na | 15 | | | | | Mg stearate | 1.2 | | | | | | | 30 |
| Atorvastatin | 10 | Mannitol | 182.6 | | | L-HPC | 15 | CC-Na | 15 | | | | | Mg stearate | 1.2 | | | | | | | 30 |
| Montelukast | 5.2 | Mannitol | 70 | MCC | 48.8 | | | PVPP | 20 | | | | | Mg stearate | 2 | Sodium Bicarbonate | 0 | 140 | 0.06 | 3.79 | | 40 |
| Montelukast | 5.2 | Mannitol | 0 | MCC | 116.8 | | | PVPP | 20 | | | | | Mg stearate | 2 | Sodium Bicarbonate | 16 | 97 | 0.11 | 3.78 | | 10 |
| Montelukast | 5.2 | Mannitol | 0 | MCC | 116.8 | | | PVPP | 6 | | | | | Mg stearate | 2 | Sodium Bicarbonate | 16 | 93 | 0.04 | 3.07 | | 15 |
| Montelukast | 5.2 | Mannitol | 0 | MCC | 92.41 | | | PVPP | 4.8 | | | | | Mg stearate | 1.596 | Sodium Bicarbonate | 12.80 | 158 | 0.17 | 3.79 | | 8 |
| Montelukast | 5.2 | Mannitol | 0 | MCC | 133.1 | | | PVPP | 6.8 | | | | | Mg stearate | 2.261 | Sodium Bicarbonate | 18.14 | 103 | 0.08 | 3.74 | | 35 |
| Montelukast | 5.2 | Mannitol | 0 | MCC | 110.8 | | | PVPP | 6 | | | | | Mg stearate | 2 | Sodium Bicarbonate | 22.01 | 85 | 0.06 | 3.77 | | 5 |
| Montelukast | 5.2 | Mannitol | 0 | MCC | 113.8 | | | PVPP | 9 | | | | | Mg stearate | 2 | Sodium Bicarbonate | 16 | 87 | 0.06 | 3.76 | | 10 |
| Amlodipine | 5 | Mannitol | 25 | MCC | 40 | | | PVPP | 40 | | | | | Mg stearate | 2 | SDS | 1 | 29 | 0.12 | 4.06 | 9 | 19.8 |
| Nisoldipine | 50 | Mannitol | 70 | MCC | 58 | PVP | 40 | CC-Na | 10 | PVPP | 10 | | | Mg stearate | 2 | | | | 0.44 | | 8 | 36 |
| Nisoldipine | 50 | Mannitol | 70 | MCC | 58 | PVP | 0 | CC-Na | 10 | PVPP | 10 | | | Mg stearate | 2 | | | | 0.67 | | 8 | 30 |
| Nisoldipine | 50 | Mannitol | 70 | MCC | 98 | | | CC-Na | 10 | PVPP | 10 | | | Mg stearate | 2 | | | | 0.52 | | 8 | 90 |
| Donepezil | 10 | Mannitol | 170.1 | | | | | CC-Na | 0 | PVPP | 56 | | | Mg stearate | 3 | | | 54 | 0.87 | 4.12 | 9.5 | 11 |
| Donepezil | 10 | Mannitol | 198.1 | | | | | CC-Na | 0 | PVPP | 28 | | | Mg stearate | 3 | | | 59 | 0.62 | 4.14 | 9.5 | 15 |
| Donepezil | 10 | Mannitol | 170.1 | | | | | CC-Na | 0 | PVPP | 0 | | | Mg stearate | 3 | | | 67 | 0.43 | 4.11 | 9.5 | 38 |
| Donepezil | 10 | Mannitol | 170.1 | | | | | CC-Na | 56 | PVPP | 0 | | | Mg stearate | 3 | | | 55 | 0.52 | 4.12 | 9.5 | 7.11 |
| Donepezil | 10 | Mannitol | 198.1 | | | | | CC-Na | 0 | PVPP | 0 | | | Mg stearate | 3 | | | 69 | 0.57 | 4.12 | 9.5 | 73 |
| Donepezil | 10 | Mannitol | 170.1 | | | | | CC-Na | 0 | PVPP | 56 | | | Mg stearate | 3 | | | 54 | 0.87 | 4.12 | 9.5 | 11 |
| Lamotrigine | 25 | Mannitol | 47.05 | | | | | PVPP | 2.5 | | | | | Mg stearate | 0.75 | | | 40 | 0.84 | | 5 | 17.21 |
| Lamotrigine | 25 | Mannitol | 44.25 | | | | | PVPP | 5 | | | | | Mg stearate | 0.75 | | | 40 | 0.27 | | 5 | 12.33 |
| Lamotrigine | 25 | Mannitol | 44.25 | | | | | PVPP | 5 | | | | | Mg stearate | 0.75 | | | 10 | 1.03 | | 5 | 3.72 |
| Lamotrigine | 25 | Mannitol | 45.25 | | | | | PVPP | 3.75 | | | | | Mg stearate | 1 | | | 10 | 1.52 | | 5 | 4.04 |
| Lamotrigine | 25 | Mannitol | 45.75 | | | | | PVPP | 3.75 | | | | | Mg stearate | 0.5 | | | 10 | 1.36 | | 5 | 3.47 |
| Lamotrigine | 25 | Mannitol | 45.25 | | | | | PVPP | 3.75 | | | | | Mg stearate | 1 | | | 40 | 0.42 | | 5 | 17.17 |

| API | Amount | Filler | Amount | Filler2 | Amount2 | Disintegrant | Amount | Lubricant | Amount | Glidant | Amount | Other | Amount |  |  |  |  |  |
|---|---|---|---|---|---|---|---|---|---|---|---|---|---|---|---|---|---|---|
| Lamotrigine | 25 | Mannitol | 56.5 | | | PVPP | 2.5 | Mg stearate | 1 | | | | | 25 | 2.1 | | 5 | 8 |
| Lamotrigine | 25 | Mannitol | 44.5 | | | PVPP | 5 | Mg stearate | 0.5 | | | | | 25 | 0.68 | | 5 | 5.85 |
| Lamotrigine | 25 | Mannitol | 45.75 | | | PVPP | 3.75 | Mg stearate | 0.5 | | | | | 40 | 0.64 | | 5 | 10.5 |
| Lamotrigine | 25 | Mannitol | 46.5 | | | PVPP | 2.5 | Mg stearate | 1 | | | | | 25 | 0.49 | | 5 | 7.5 |
| Lamotrigine | 25 | Mannitol | 45.5 | | | PVPP | 3.75 | Mg stearate | 0.75 | | | | | 25 | 0.46 | | 5 | 6.12 |
| Lamotrigine | 25 | Mannitol | 46.5 | | | PVPP | 2.5 | Mg stearate | 0.75 | | | | | 10 | 1.7 | | 5 | 3.99 |
| Lamotrigine | 25 | Mannitol | 45.5 | | | PVPP | 3.75 | Mg stearate | 0.75 | | | | | 25 | 0.51 | | 5 | 8.1 |
| Clozapine | 12.5 | Mannitol | 94.6 | MCC | 21 | CMS-Na | 11.2 | Mg stearate | 0.7 | | | | | 31 | 0.68 | | 8.5 | 14.6 |
| Clozapine | 12.5 | Mannitol | 39.9 | MCC | 21 | CMS-Na | 11.2 | Mg stearate | 0.7 | | | 2-hydroxypropyl-β-cyclodextrin | 54.7 | 32 | 0.63 | | 8.5 | 15.3 |
| Tramadol | 50 | Mannitol | 172 | | | PVPP | 18 | Mg stearate | 2 | Aerosil | 10 | | | 31 | 0.55 | | 9 | 47 |
| Tramadol | 50 | Mannitol | 166 | | | PVPP | 18 | Mg stearate | 2 | Aerosil | 10 | | | 32 | 0.69 | | 9 | 34 |
| Tramadol | 50 | Mannitol | 172 | | | PVPP | 12 | Mg stearate | 2 | Aerosil | 2 | | | 34 | 0.58 | | 9 | 72 |
| Tramadol | 50 | Mannitol | 178 | | | PVPP | 18 | Mg stearate | 2 | Aerosil | 2 | | | 33 | 0.62 | | 9 | 61 |
| Sildenafil | 29.8 | Mannitol | 251.2 | | | PVPP | 13 | Mg stearate | 3 | Aerosil | 0 | | | 52 | 0.19 | | 10 | 25 |
| Sildenafil | 29.8 | Mannitol | 236.2 | | | PVPP | 13 | Mg stearate | 3 | Aerosil | 0 | | | 40 | 0.3 | | 10 | 26 |
| Sildenafil | 29.8 | Mannitol | 221.2 | | | PVPP | 13 | Mg stearate | 3 | Aerosil | 0 | | | 35 | 0.41 | | 10 | 25 |
| Sildenafil | 29.8 | Mannitol | 206.2 | | | PVPP | 13 | Mg stearate | 3 | Aerosil | 0 | | | 30 | 0.49 | | 10 | 26 |
| Sildenafil | 29.8 | Mannitol | 205.5 | | | PVPP | 13 | Mg stearate | 3 | Aerosil | 0.75 | | | 32 | 0.46 | | 10 | 27 |
| Sildenafil | 29.8 | Mannitol | 204.7 | | | PVPP | 13 | Mg stearate | 3 | Aerosil | 1.5 | | | 35 | 0.33 | | 10 | 26 |
| Sildenafil | 29.8 | Mannitol | 203.5 | | | PVPP | 13 | Mg stearate | 3 | Aerosil | 2.25 | | | 33 | 0.3 | | 10 | 27 |
| Ondansetron | 8 | Mannitol | 22.5 | MCC | 18.87 | PVPP | 3.75 | Aerosil | 0.75 | | | | | 29 | 0.44 | 2.6 | 5.5 | 8.53 |
| Ondansetron | 8 | Mannitol | 22.5 | MCC | 15.12 | PVPP | 7.5 | Aerosil | 0.75 | | | | | 22 | 0.38 | 2.57 | 5.5 | 10.17 |
| Ondansetron | 8 | Mannitol | 22.5 | MCC | 11.37 | PVPP | 11.25 | Aerosil | 0.75 | | | | | 27 | 0.4 | 2.77 | 5.5 | 7.33 |
| Ondansetron | 8 | Mannitol | 22.5 | MCC | 18.87 | PVPP | 3.75 | Aerosil | 0.75 | | | | | 26 | 0.53 | 2.72 | 5.5 | 6 |
| Ondansetron | 8 | Mannitol | 22.5 | MCC | 15.12 | PVPP | 7.5 | Aerosil | 0.75 | | | | | 25 | 0.48 | 2.74 | 5.5 | 11.17 |
| Ondansetron | 8 | Mannitol | 22.5 | MCC | 11.37 | PVPP | 11.25 | Aerosil | 0.75 | | | | | 23 | 0.59 | 2.83 | 5.5 | 7.17 |

| Drug | Dose | Diluent | | Binder | | Disintegrant | | Glidant | | Lubricant | | Hardness | Friability | Thickness | Diameter | DT |
|---|---|---|---|---|---|---|---|---|---|---|---|---|---|---|---|---|
| Ondansetron | 8 | Mannitol | 22.5 | MCC | 18.87 | PVPP | 3.75 | Aerosil | 0.75 | | | 25 | 0.37 | 2.67 | 5.5 | 7 |
| Ondansetron | 8 | Mannitol | 22.5 | MCC | 18.87 | PVPP | 3.75 | Aerosil | 0.75 | | | 28 | 0.49 | 2.68 | 5.5 | 28.5 |
| Ondansetron | 8 | Mannitol | 22.5 | MCC | 15.12 | PVPP | 7.5 | Aerosil | 0.75 | | | 24 | 0.54 | 2.67 | 5.5 | 16.33 |
| Ondansetron | 8 | Mannitol | 22.5 | MCC | 11.37 | PVPP | 11.25 | Aerosil | 0.75 | | | 24 | 0.62 | 2.66 | 5.5 | 26 |
| Ondansetron | 8 | Mannitol | 22.5 | MCC | 21.87 | PVPP | 0.75 | Aerosil | 0.75 | | | 23 | 0.55 | 2.54 | 5.5 | 33 |
| Ondansetron | 8 | Mannitol | 22.5 | MCC | 20.37 | PVPP | 2.25 | Aerosil | 0.75 | | | 25 | 0.46 | 2.46 | 5.5 | 21.17 |
| Ondansetron | 8 | Mannitol | 22.5 | MCC | 18.87 | PVPP | 3.75 | Aerosil | 0.75 | | | 25 | 0.39 | 2.7 | 5.5 | 15.33 |
| Ondansetron | 8 | Mannitol | 22.5 | MCC | 18.87 | PVPP | 3.75 | Aerosil | 0.75 | | | 27 | 0.29 | 2.54 | 5.5 | 15.67 |
| Ondansetron | 8 | Mannitol | 22.5 | MCC | 17 | PVPP | 5.63 | Aerosil | 0.75 | | | 26 | 0.33 | 2.62 | 5.5 | 13.67 |
| Diclofenac sodium | 50 | MCC | 10 | Lactose | 131 | | | Aerosil | 5 | Mg stearate | 4 | 55 | 0.68 | | | 8.5 |
| Fenoverine | 100 | Mannitol | 93.75 | MCC | 37.5 | CC-Na | 10 | Aerosil | 2.5 | Mg stearate | 1.25 | 30 | | 5.5 | 8 | 70 |
| Fenoverine | 100 | Mannitol | 88.75 | MCC | 37.5 | CC-Na | 15 | Aerosil | 2.5 | Mg stearate | 1.25 | 27 | | 5.5 | 8 | 55 |
| Fenoverine | 100 | Mannitol | 83.75 | MCC | 37.5 | CC-Na | 20 | Aerosil | 2.5 | Mg stearate | 1.25 | 25 | | 5.5 | 8 | 40 |
| Fenoverine | 100 | Mannitol | 93.75 | MCC | 37.5 | PVPP | 10 | Aerosil | 2.5 | Mg stearate | 1.25 | 24 | | 5.5 | 8 | 21 |
| Fenoverine | 100 | Mannitol | 88.75 | MCC | 37.5 | PVPP | 15 | Aerosil | 2.5 | Mg stearate | 1.25 | 24 | | 5.6 | 8 | 19 |
| Fenoverine | 100 | Mannitol | 83.75 | MCC | 37.5 | PVPP | 20 | Aerosil | 2.5 | Mg stearate | 1.25 | 23 | | 5.5 | 8 | 18 |
| Fenoverine | 100 | Mannitol | 93.75 | MCC | 37.5 | CMS-Na | 10 | Aerosil | 2.5 | Mg stearate | 1.25 | 25 | | 5.6 | 8 | 37 |
| Fenoverine | 100 | Mannitol | 88.75 | MCC | 37.5 | CMS-Na | 15 | Aerosil | 2.5 | Mg stearate | 1.25 | 26 | | 5.4 | 8 | 30 |
| Fenoverine | 100 | Mannitol | 83.75 | MCC | 37.5 | CMS-Na | 20 | Aerosil | 2.5 | Mg stearate | 1.25 | 25 | | 5.6 | 8 | 31 |

Table 2 The accuracies of OFDT on training, testing, and final testing sets

| Network | Training Set (%) | Validation Set (%) | Testing Set (%) |
|---|---|---|---|
| ANN | 85.60 | 80.00 | 75.00 |
| DNN | 85.60 | 85.00 | 80.00 |

**Declaration of interest**

The authors report no conflicts of interest. The authors alone are responsible for the content and writing of this article.